\documentclass{article}
\usepackage{spconf,amsmath,graphicx}
\usepackage{tabularx}
\usepackage{booktabs}
\usepackage{textcomp}
\usepackage{stfloats}
\usepackage{url}
\usepackage{verbatim}
\usepackage{graphicx}
\usepackage{cite}
\usepackage{multirow}
\usepackage{threeparttable}
\usepackage{tabularx}
\usepackage[usenames,dvipsnames]{color}
\usepackage{enumitem}
\usepackage{bbding}
\usepackage{amsmath,amsfonts}
\usepackage{algorithmic}
\usepackage{algorithm}
\usepackage{array}
\newcolumntype{Y}{>{\centering\arraybackslash}X}
\newcolumntype{Z}{>{\arraybackslash}X}

\definecolor{Color1}{RGB}{180,199,231}
\definecolor{Color2}{RGB}{236,140,211}

\allowdisplaybreaks[4]

\usepackage{makecell}
\usepackage{subfig}
\usepackage{float}
\usepackage{hyperref}

\title{A New People-Object Interaction Dataset and NVS Benchmarks}
\name{\begin{tabular}{c}
Shuai Guo$^{1}$, Houqiang Zhong$^{1}$, Qiuwen Wang$^{1}$, Ziyu Chen$^{1}$, Yijie Gao$^1$, Jiajing Yuan$^{2}$, Chenyu Zhang$^{3}$,\\Rong Xie$^{1}$, Li Song$^{1,4}$\sthanks{Corresponding Author. (e-mail: song\_li@sjtu.edu.cn)}\thanks{This work was supported by the Fundamental Research Funds for the Central Universities, National Key R\&D Project of China (2019YFB1802701), MoE-China Mobile Research Fund Project (MCM20180702), STCSM under Grant 22DZ2229005, 111 project BP0719010.}
\end{tabular}}

\address{
\begin{tabular}{c}
$^{1}$ Institute of Image Communication and Network Engineering, Shanghai Jiao Tong University,\\Shanghai, China \\
$^{2}$ School of Electronic Information and Electrical Engineering, Shanghai Jiao Tong University,\\Shanghai, China\\
$^{3}$ SJTU-ParisTech Elite Institute of Technology, Shanghai Jiao Tong University, Shanghai, China\\
$^{4}$ Cooperative Medianet Innovation Center, Shanghai Jiao Tong University, Shanghai, China
\end{tabular}
}

\begin{document}
%
\maketitle
\begin{abstract}
Recently, NVS in human-object interaction scenes has received increasing attention.
Existing human-object interaction datasets mainly consist of static data with limited views, offering only RGB images or videos, mostly containing interactions between a single person and objects.
Moreover, these datasets exhibit complexities in lighting environments, poor synchronization, and low resolution, hindering high-quality human-object interaction studies.
In this paper, we introduce a new people-object interaction dataset that comprises 38 series of 30-view multi-person or single-person RGB-D video sequences, accompanied by camera parameters, foreground masks, SMPL models, some point clouds, and mesh files.
Video sequences are captured by 30 Kinect Azures, uniformly surrounding the scene, each in 4K resolution 25 FPS, and lasting for 1$\sim$19 seconds.
Meanwhile, we evaluate some SOTA NVS models on our dataset to establish the NVS benchmarks.
We hope our work can inspire further research in human-object interaction.
\end{abstract}
\begin{keywords} 
Dataset, People-Object Interaction, Prior Information, Novel View Synthesis, Benchmark
\end{keywords}

\section{Introduction}
Human-object interaction refers to the interactions of single or multiple people with objects and is a common type of scenario in everyday life~\cite{wang2023hierarchical}.
In recent years, research on novel view synthesis (NVS) on human-object interaction has received much attention.
NVS allows for the synthetic generation of a virtual view placed at an arbitrarily selected position in a three-dimensional scene.
NVS in scenes of human-object interaction is necessary for high-level vision tasks like action analysis, visual scene answering, and video understanding.
The main challenges are the complex interaction patterns and the severe occlusions~\cite{su2022robustfusion}.

However, existing human-object interaction datasets mainly consist of static data with limited viewpoints, offering only RGB images or videos, mostly containing interactions between a single person and objects.
Most existing datasets have complex lighting environments, poor synchronization, and low resolution.
Some of these datasets only focus on hand-object interactions.
These problems make it difficult to conduct high-quality dynamic interaction studies and impede progress in addressing the challenges of NVS research of human-object interaction.

To address the aforementioned issues and assist researchers in tackling the challenges of human-object interaction NVS studies, this paper introduces a people-object interaction dataset that offers richer prior information, more viewpoints, and higher-quality video sequences.
Our people-object interaction dataset contains 38 series of RGB-D video sequences of a single person or multiple people that interact with objects.
We also provide the corresponding camera parameters, foreground mask, skinned multi-person linear (SMPL) models, some point clouds, and mesh files.
Our dataset is publicly available under the GPL-3.0 license\footnote{\href{https://github.com/sjtu-medialab/People-Ojbect-Interaction-Dataset}{https://github.com/sjtu-medialab/People-Ojbect-Interaction-Dataset}}.

The RGB-D videos are captured by identical 30 Kinect Azures that evenly surround the scene.
Each video sequence is 4K, 25 FPS, 1$\sim$19 seconds in duration.
The foreground masks, point clouds, SMPL models, and mesh files are obtained by algorithmic post-processing.
Multiple viewpoints, depth sequences, foreground masks, point clouds, SMPL models, and mesh files can be used as training data or prior inputs for the people-object interaction model.
They provide rich prior information that can effectively character the interaction mode of the human-object interaction at different levels and alleviate the reconstruction difficulties caused by complex overlaps.

Meanwhile, we conduct experiments with some state-of-the-art (SOTA) NVS methods in our dataset to obtain the NVS benchmarks.
The experimental results show that existing methods generally perform well on the training set but very poorly on the test set.
Our dataset uncovers the substantial overfitting issues that may present in current methods within the field of human-object interaction NVS research, further emphasizing the critical importance of our dataset.

The main contributions of this paper include 1) a new multiplexed synchronized RGB-D people-object interaction dataset, and its corresponding camera parameters, foreground masks, SMPL models and some point clouds, mesh files, and 2) benchmarks of the NVS on our dataset.


\section{Related Work} \label{relatedwork}
In this section, we review some usually used datasets in the research of human-object interaction, including static human-object interaction datasets and dynamic human-object interaction datasets.

\subsection{Static Human-object Interaction Dataset}
RICH dataset~\cite{huang2022capturing} is a comprehensive dataset that contains multi-view 4K video sequences, 3D human ground truth, 3D human scans, and 3D scene scans, for studying human interaction and contact in real scenes.
Another dataset that focuses on human body expressions is HUMBI~\cite{yoon2021humbi}, which is a large-scale multiview dataset that contains view-specific appearance and geometry information of eye gaze, face, hand, body, and clothing of 772 different people.
For single-person pose estimation, the MPII Human Pose dataset~\cite{andriluka20142d} is a widely used dataset that consists of about 25K images from YouTube, each with manual annotations of up to 16 body joints.
Moreover, COCO dataset~\cite{chen2015microsoft} is a popular dataset that mainly aims at object detection and segmentation, which contains more than 330K images with 1.5 million targets and multiple target and material categories, each with a five-sentence description of the image.
PaStaNet~\cite{li2020pastanet} is a dataset for fine-grained action recognition of human-object interactions, which contains about 10K video clips from YouTube, each with annotations of human pose, object category, and action category.

\subsection{Dynamic Human-object Interaction Dataset}
Something-Something V2 dataset~\cite{goyal2017something} contains 220,847 video clips of humans performing basic actions with common objects, each with a label.
JHMDB dataset~\cite{jhuang2013towards} shows 51 human actions with multiple annotations.
For human actions in indoor scenes, the Charades dataset~\cite{sigurdsson2016hollywood} is a dataset of videos that are filmed by crowd workers following scripts, each video has multiple annotations.
D3D-HOI dataset~\cite{xu2021d3d} consists of monocular videos that show human interactions with articulated objects captured from different scenes and viewpoints, each video has ground truth annotations of 3D object pose, shape, and partial motion.
EgoHOS~\cite{zhang2022fine} is a dataset of egocentric images that show fine-grained hand-object segmentation for human-object interactions in daily activities, each image has pixel-level labels for the hand and the object.
For action recognition and understanding in first-person perspectives, EPIC-KITCHENS dataset~\cite{damen2020rescaling} encompasses egocentric video recordings in kitchen environments, offering a rich resource for action recognition and understanding daily activities in first-person perspectives.
NTU RGB+D 120 dataset~\cite{liu2019ntu} is a large-scale multiview dataset of human body expressions, containing view-specific appearance and geometry information of eye gaze, face, hand, body, and clothing of 120 different people.
ARCTIC dataset~\cite{fan2023arctic} consists of RGB-D video sequences of human cooking activities recorded by the Microsoft Kinect sensor, containing 4 subjects, 12 activities, and 120 videos.
It also provides hand-object contact information.

\section{Dataset Details} \label{datasetdetails}
In this section, we provide detailed explanations of our sequences collection, camera calibration, and post-processing procedures, followed by a comparative discussion with similar datasets.


\begin{figure}
\centering
\subfloat[Our collection system.]{\includegraphics[width=.625\columnwidth]{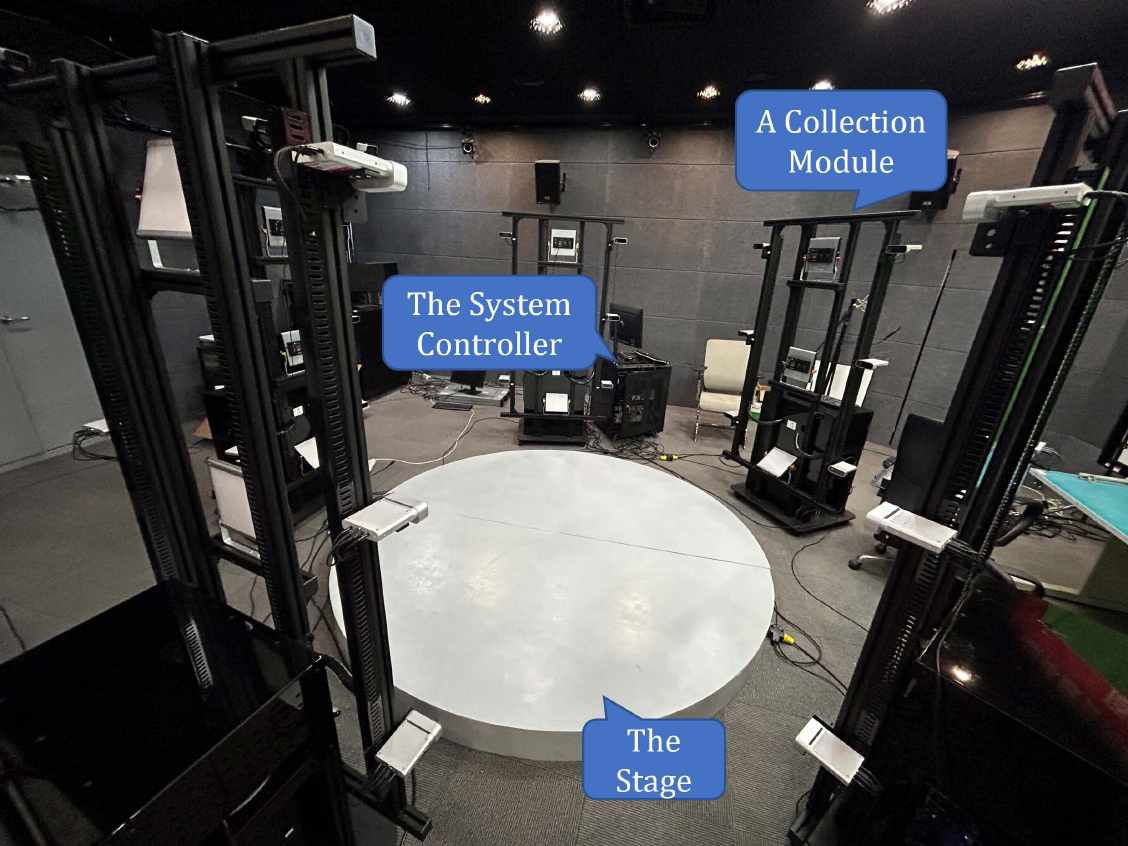}}\hspace{2pt}
\subfloat[A collection module.]{\includegraphics[width=.35\columnwidth]{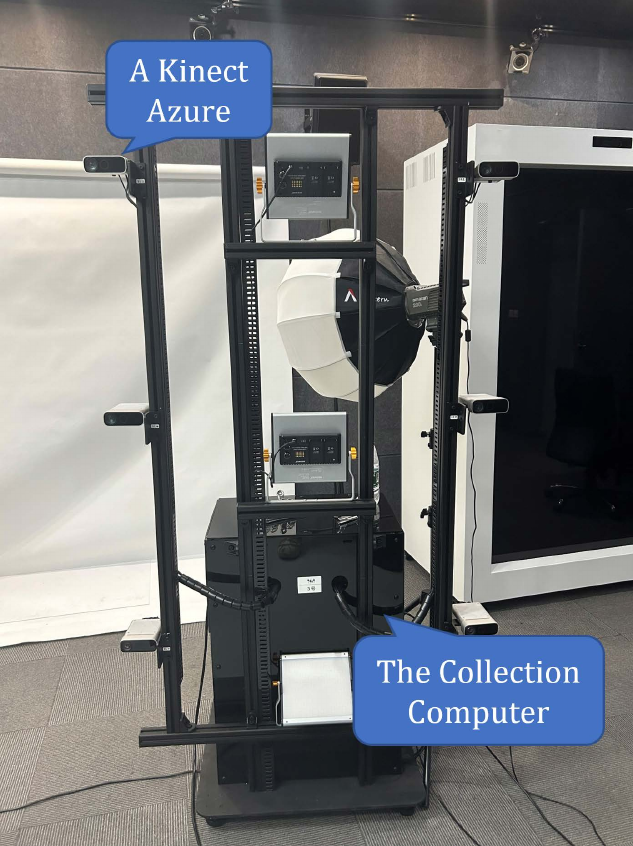}}\\
\caption{Our collection system includes 5 collection modules and 1 system controller.}
\label{system}
\end{figure}

\subsection{Sequences Collection}
To collect the video sequences, as shown in Fig.~\ref{system}, we develop a new system that consists of 5 collection modules and 1 system controller.
The 5 collection modules are evenly spaced around a circular stage with a diameter of 2.5m.
Each module is equipped with 1 collection computer and 6 Kinect Azures, arranged in 3 rows and 2 columns, with each row about 0.65 meters apart and each column about 0.8 meters apart.
The Kinect Azures belonging to the same module are connected in series.
During the collection of video sequences, the controller sends synchronization signals to the collection modules.
The collection modules then send synchronization signals to the Kinect Azures they contain, ensuring synchronous power-on and collection of all 30 Kinect Azures.
RGB and Depth images are both collected at a resolution of 4K (3840$\times$2160).
Depth images are aligned with RGB images with Azure Kinect SDK.

\begin{table*}
\caption{Details of The Video Sequences in Our People-Object Interaction Dataset.}
\centering
\begin{tabularx}{\linewidth}{p{4cm}<{\centering}p{1cm}<{\centering}p{2cm}<{\centering}Z}
\cmidrule{1-4}
Category & Amount & Duration (s) & \qquad \qquad \qquad \qquad \qquad \quad Contents\\
\cmidrule{1-4}
Empty Scene & 1 & 1 & The empty scene that has nobody on the stage.\\
Camera Calibration & 1 & 8 & Camera calibration sequences.\\
One Person with Objects & 23 & 2$\sim$19 & Flipping through a book, circling a chair before sitting down, opening and closing an umbrella, pushing a suitcase, putting on a safety helmet, typing on a laptop, and so on. \\
Two People with Objects & 11 & 2$\sim$14 & Two people working together to move a table; two people collaborating to sweep the floor; two people hurrying along, one carrying a backpack and the other pulling a suitcase, two people holding a chessboard, and so on. \\
Three People with Objects & 2 & 2$\sim$5 & Three people taking a group photo together, three people taking pictures of each other.\\
\cmidrule{1-4}
\end{tabularx}
\label{contents}
\end{table*}

Afterward, we invite some volunteers to perform people-object interactions on the circular stage and capture the sequences with our collection system.
The interacting objects are common items from everyday life, such as laptops, tables, chairs, suitcases, mobile phones, water cups, backpacks, and so on.
Then, we compress all the captured contents into 4K (3840$\times$2160), 25 FPS video sequences using HEVC codec.
To facilitate researchers in conducting further studies, our dataset provides both video sequences of empty scenes and camera calibration.
Details of the video sequences of our dataset are shown in Table~\ref{contents}.

\subsection{Camera Calibration}
\begin{figure}
\centering
\subfloat[Our chessboard and the calibration of two collection modules.]{\includegraphics[width=.49\columnwidth]{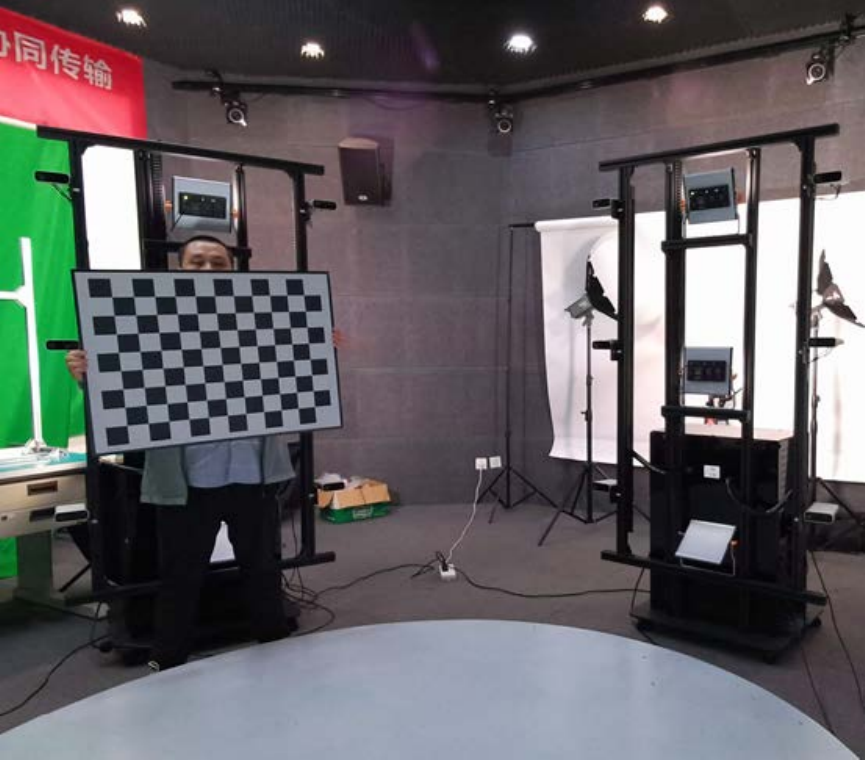}}\hspace{2pt}
\subfloat[Visualization of the calibrated world positions of Kinect Azures.]{\includegraphics[width=.49\columnwidth]{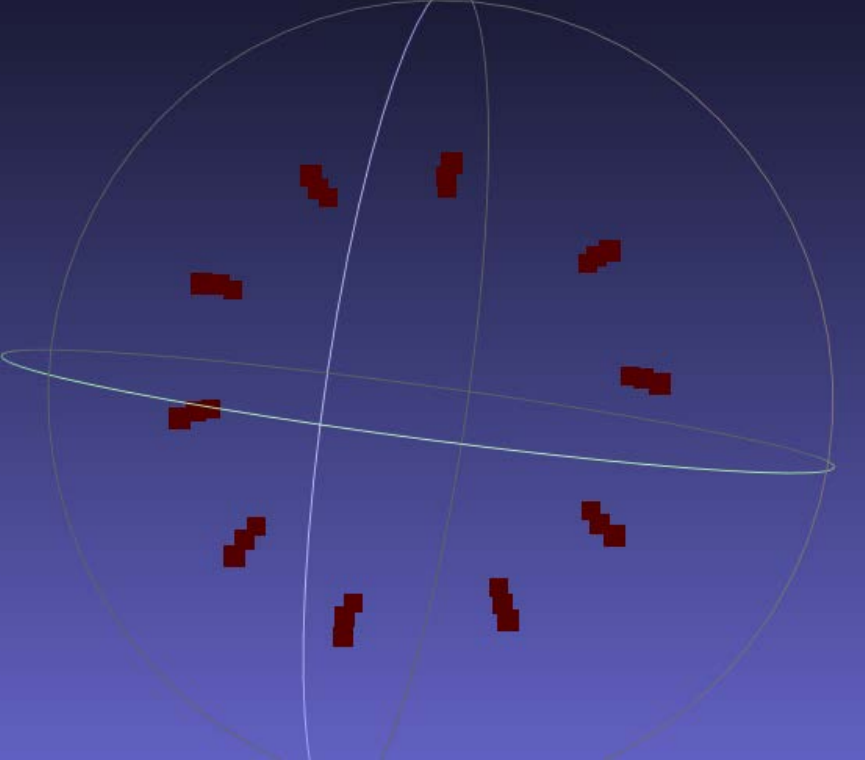}}\\
\caption{Calibration of the 30 Kinect Azures and visualization of their calibrated world positions, which essentially match the actual distributions of them.}
\label{calibration}
\end{figure}

Since the 30 Kinect Azures are arranged surrounding the scene, no chessboard can be simultaneously visible to all of them. 
Therefore, we first perform a calibration for 12 Kinect Azures of any two neighboring collection modules with a chessboard and OpenCV.
Then we transfer the extrinsic parameters of all 30 Kinect Azure to the same world coordinate system.
Fig.~\ref{calibration}. (a) shows our chessboard and the calibration of two collection modules.

Assuming that there are three neighbouring collection modules $1$, $2$, and $3$, we have performed two calibrations for modules $1$,$2$ and $2$,$3$, respectively.
Then each of the two calibrations has obtained a world coordinate system.
Since the 6 Kinect Azures on the collection module $2$ are calibrated twice, we denote their world coordinates on the two world coordinate systems as $X_W$ and $X_W^{'}$, respectively. 
Then $X_W$ and $X_W^{'}$ satisfy the following equations:
\begin{align}
\begin{split}
X_{c}^{1}&=K^{1}(R^{1}X_{W}+T^{1}),\\
X_{c}^{2}&=K^{2}(R^{2}X_{W}+T^{2}),\\
X_{c}^{2'}&=K^{2'}(R^{2'}X_{W}'+T^{2'}),\\
X_{c}^{3'}&=K^{3'}(R^{3'}X_{W}'+T^{3'}),\\
\end{split}
\end{align}
where $K$, $R$, and $T$ represent the intrinsic parameters, rotation matrix, and translation matrix obtained from the calibration process.
The superscripts $1$, $2$, $3$ indicate collection model $1$, $2$, and $3$, respectively.
The subscript $W$ denotes world coordinates, the subscript $c$ denotes the camera coordinates.

As $X_W$ and $X_W^{'}$ represent the 6 identical points in different world coordinate systems, there must exist an affine transformation: 
\begin{equation}
X_W' = rX_W + t,
\end{equation}
where $r$ and $t$ denote the rotation and translation components of the affine transformation, respectively.

We use the Kabsch algorithm~\cite{Kabsch} to solve the affine transformation to obtain $r$ and $t$.
The Kabsch algorithm is a method for determining the optimal rotation matrix based on two sets of corresponding points, minimizing the mean squared distance between the point sets.
This algorithm is usually used for molecular alignment and point alignment in computer graphics.
Therefore, we have:
\begin{equation}
X_{c}^{3'} = K^{3'}[(R^{3'}r)X_{W}+(t+T^{3'})].
\end{equation}

With this equation, for any point located in the world coordinate system calibrated by collection modules $1$,$2$ and observed by collection module $3$, we can obtain its camera coordinates that correspond to the Kinect Azures on collection module $3$.
Thus, the transformation of the extrinsic parameters of the Kinect Azures on collection module $3$ to the world coordinate system calibrated by collection modules $1$,$2$ is accomplished.

Similarly, we transfer the extrinsic parameters of all Kinect Azures to the world coordinate system calibrated by collection modules $1$,$2$, thus completing the camera calibration.
Fig.~\ref{calibration}. (b) shows the visualization of world positions of Kinect Azures based on the calibration results.
It can be seen that the world coordinates of the Kinect Azures obtained from our calibration closely match their actual distribution, indicating that our calibration is essentially accurate.

\begin{figure*}
\centering
\includegraphics[width = \linewidth]{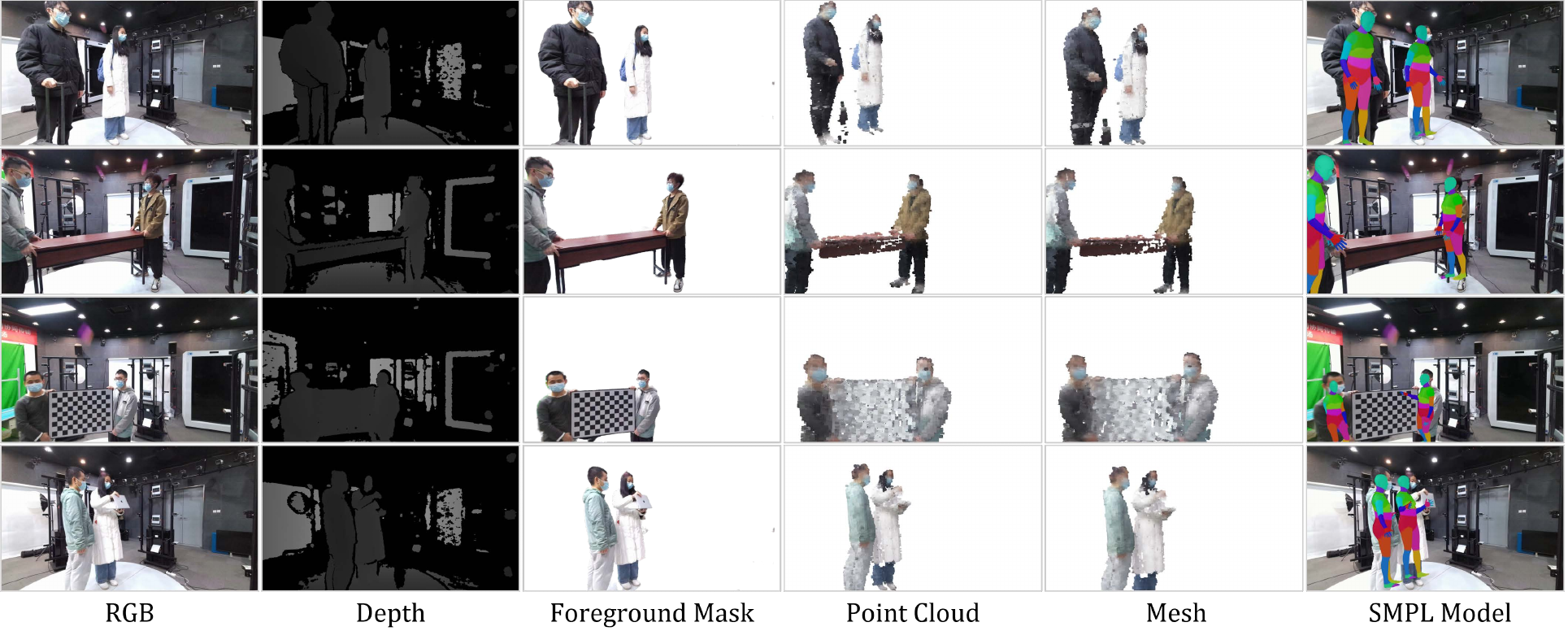}
\caption
{Some RGB-D frames and their corresponding foreground masks, point clouds, mesh files, and SMPL models in our people-object interaction dataset.}
\label{frames}
\end{figure*}

\subsection{Post processes}
To obtain foreground masks, point clouds, mesh files, and SMPL models, we conducted post-processing on the collected video sequences.
For each video sequence, we obtain a foreground mask sequence in 4K, 25 FPS.
Point clouds, mesh files, and SMPL models are obtained for some selected frames of our dataset.
Fig.~\ref{frames} shows some RGB-D frames and their corresponding foreground masks, point clouds, mesh files, and SMPL models in our dataset.
The algorithms we used are detailed below.

\noindent \textbf{Foreground masks.}
We use the real-time high-resolution background matting method introduced by Lin et al.~\cite{lin2021real} and the captured empty scene to obtain foreground masks.
Real-time high-resolution background matting employs a base network that computes a low-resolution result which is refined by a second network operating at high resolution on selective patches.
It efficiently refines only the error-prone regions at high resolution, requiring an additional background frame to be captured and used to recover the alpha matte and the foreground layer.

\noindent \textbf{Point clouds.}
We use the method of Zhou et al.~\cite{zhouibc} to obtain the point clouds.
With camera parameters and RGB-D sequences, for each frame, we first map pixels of each view to the world coordinates to obtain a sparse point cloud for each view.
Iterative closest points (ICP) are used to reduce the matching error.
Then we remove the overlapping regions of each view, which are determined with forward-backward consistency.
Afterward, a Step Discontinuity Constrained (SDC) filter is used to remove the noises, the missing pixels, and the unstable pixels.

\noindent \textbf{Mesh files.}
After the construction of point clouds, the implicitly stored isosurface is extracted through ray casting.
A ray is projected from the start to the endpoint, intersecting a sequence of voxels along its path.
As the ray progresses, the Truncated Signed Distance Function (TSDF) of each voxel it crosses is evaluated to pinpoint the surface interface, or the zero-crossing point, where the TSDF values shift from positive to negative in the marching direction.
The mesh is then constructed by connecting these adjacent zero-crossing points~\cite{zhou2022rgbd}.

\noindent \textbf{SMPL models.}
We use the MMHuman3D library~\cite{mmhuman3d} for the extractions of SMPL models.
MMHuman3D is an open-source PyTorch-based codebase for the use of 3D human parametric models in computer vision and computer graphics.
MMHuman3D reimplements popular methods, allowing users to reproduce SOTAs with one line of code.
It provides a demo script to estimate SMPL parameters for single-person or multi-person from the input image or video.
We use the models pre-trained by MMHuman3D to get the estimated SMPL models.

\begin{table*}
\caption{Comparisons Between Our Dataset and Usually Used Human-Object Interaction Datasets}
\centering
\begin{tabularx}{\textwidth}{p{1.8cm}<{\centering}YYp{1.8cm}<{\centering}YYYYYY}
\cmidrule{1-10}
Dataset & Series & View & Resolution & Dynamic & Depth & M.P. & P.C. & Mesh & SMPL\\
\cmidrule{1-10}
MPII~\cite{andriluka20142d} & 823 & 1 & 640$\times$480 & \XSolidBrush & \XSolidBrush & \XSolidBrush & \XSolidBrush & \XSolidBrush &\XSolidBrush\\
PaStaNet~\cite{li2020pastanet} & 156 & 1 & 500$\times$500 & \XSolidBrush & \XSolidBrush & \Checkmark & \Checkmark & \XSolidBrush & \XSolidBrush\\
RICH~\cite{huang2022capturing} & 142 & 6-8 & 3840$\times$2160 & \XSolidBrush & \XSolidBrush & \Checkmark & \XSolidBrush & \Checkmark & \Checkmark\\
HUMBI~\cite{yoon2021humbi} & 772 & 107 & 1920$\times$1080 & \XSolidBrush & \XSolidBrush & \XSolidBrush & \Checkmark & \Checkmark & \Checkmark\\
JHMDB~\cite{jhuang2013towards} & 5100 & 1$\sim$5 & 720$\times$480 & \Checkmark & \XSolidBrush & \Checkmark & \XSolidBrush & \XSolidBrush & \XSolidBrush\\
Charades~\cite{sigurdsson2016hollywood} & 9848 & 1 & 1280$\times$720 & \Checkmark & \XSolidBrush & \Checkmark & \XSolidBrush & \XSolidBrush & \XSolidBrush\\
D3D~\cite{xu2021d3d} & 256 & 2$\sim$8 & 1280$\times$720 & \Checkmark & \XSolidBrush & \Checkmark & \XSolidBrush & \Checkmark & \Checkmark\\
EgoHOS~\cite{zhang2022fine} & 8 & 1$\sim$4 & 1920$\times$1080 & \Checkmark & \XSolidBrush & \XSolidBrush & \XSolidBrush & \Checkmark & \XSolidBrush \\
E.-K.~\cite{damen2020rescaling} & 700 & 2 & 1920$\times$1080 & \Checkmark & \XSolidBrush & \XSolidBrush & \XSolidBrush & \XSolidBrush & \XSolidBrush\\
NTU~\cite{liu2019ntu} & 120 & 3 & 1920$\times$1080 & \Checkmark & \Checkmark & \Checkmark & \Checkmark & \XSolidBrush & \XSolidBrush\\
ARCTIC~\cite{fan2023arctic} & 12 & 4 & 1920$\times$1080 & \Checkmark & \Checkmark & \XSolidBrush & \Checkmark & \XSolidBrush & \Checkmark \\
\cmidrule{1-10}
Ours & 38 & 30 & 3840$\times$2160 & \Checkmark & \Checkmark & \Checkmark & \Checkmark & \Checkmark & \Checkmark\\
\cmidrule{1-10}
\end{tabularx}
\begin{tablenotes}
\item * The number of series of videos or images is indicated by Series, the number of view per series is indicated by View, multi-person interaction with objects is indicated by M.P., and point cloud is indicated by P.C. The names of some datasets are abbreviated.
\end{tablenotes}
\label{comparison}
\end{table*}

\subsection{Comparisons With Related Datasets}
In Table~\ref{comparison}, several important properties are compared between our dataset and the frequently used datasets in human-object interaction research.
As seen from this table, our dataset is the only multi-view dynamic synchronous 4K RGB-D dataset that provides interactions between multiple people and objects.
We also provide strong prior information like foreground mask, point cloud, mesh, and SMPL model.
High-resolution, high-frame-rate, multi-view video sequences are beneficial for conducting high-quality NVS research in human-object interaction scenarios.
The abundance of prior information, such as foreground masks, point clouds, meshes, and SMPL models, helps mitigate the adverse effects of severe occlusions and complex interaction patterns, thereby enhancing the quality and speed of NVS synthesis, as well as improving performance in dynamic NVS, sparse NVS, and other related aspects.

\section{NVS Benchmarks} \label{nvsbemchmarks}
In this section, we first introduce the methods, metrics, and implementation details for the NVS benchmarks in our dataset.
Then we show some qualitative and quantitative results of the methods used, along with discussions.

\subsection{Experimental Setup}

\textbf{Metrics.}
We employ PSNR, SSIM, and LPIPS to assess the quality of the predicted images.
Peak Signal-to-Noise Ratio (PSNR$\uparrow$ in dB) is used to assess the RGB reconstruction quality, with higher values being more desirable.
Structural Similarity Index (SSIM$\uparrow$ in \%) measures the potential decline in reconstructed image quality; again, higher is preferable.
Learned Perceptual Image Patch Similarity (LPIPS$\downarrow$) assesses the similarity between image patches, with lower values indicating greater similarity.

\noindent \textbf{Methods.}
We evaluate the performance of some SOTA NVS models in our dataset, including NVS methods based on NeRF like TensoRF~\cite{chen2022tensorf}, K-Planes~\cite{fridovich2023k}, and NVS method based on 3D Gaussian Splatting (3DGS)~\cite{kerbl20233d}.
It is noteworthy that all results of these methods are obtained through publicly accessible codes and standard parameter configuration.

\noindent \textbf{Implementation Details.}
In our dataset, the 5$^{th}$ frame of scene 20231205105936, the 10$^{th}$ frame of scene 2023120419-4620, and the 10$^{th}$ frame of scene 20231204201726 are chosen as scene $1$, $2$, and $3$.
All experiments are performed on an Ubuntu 20.04 server with an NVIDIA GeForce RTX 4090 Graphics Card that has 24 GB of memory.
We use every 12$^{th}$ view of the foreground mask for testing and the other views of foreground masks for training.

\begin{figure*}
\centering
\includegraphics[width = \linewidth]{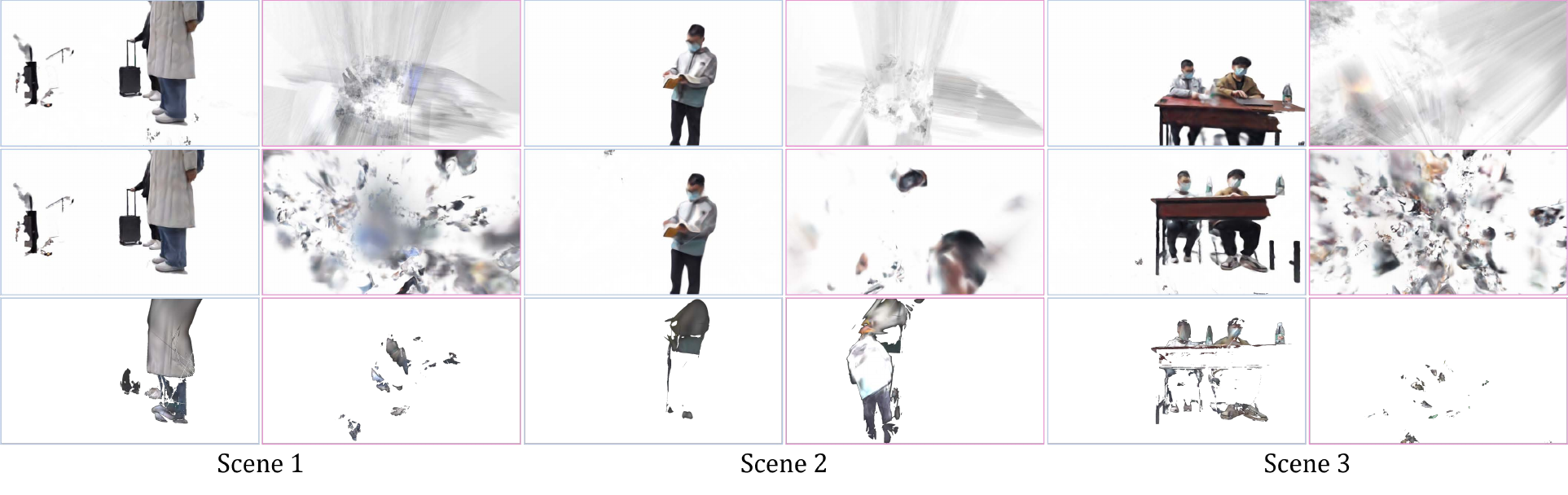}
\caption
{Qualitative results of SOTA methods on the training sets and test sets of the three scenes of our dataset. The first row shows the results of TensoRF~\cite{chen2022tensorf}, the second row shows the results of K-Planes~\cite{fridovich2023k}, the third row shows the results of 3DGS~\cite{kerbl20233d}. Results within the \textcolor{Color1}{blue borders} are from the training sets, while results within the \textcolor{Color2}{red borders} are from the test sets.}
\label{results}
\end{figure*}

\begin{table}
\caption{Average Performance of SOTA Methods on The Training Sets of The Three Scenes}
\label{traintable}
\centering
\begin{tabularx}{\linewidth}{m{2.5cm}<{\centering}YYY}
\cmidrule{1-4}
\multirow{2}*{Method} & \multicolumn{3}{c}{Metrics} \\
\cmidrule(l{0.13cm}){2-4}
~ & PSNR$\uparrow$ & SSIM$\uparrow$ & LPIPS$\downarrow$ \\
\cmidrule{1-4}
TensoRF~\cite{chen2022tensorf} & 33.47 & 0.97 & 0.08 \\
K-Planes~\cite{fridovich2023k} & 28.50 & 0.95 & 0.11 \\
3DGS~\cite{kerbl20233d} & 22.80 & 0.93 & 0.08 \\
\cmidrule{1-4}
\end{tabularx}
\end{table}

\subsection{Experimental Results}
In Fig.~\ref{results}, we show some qualitative results of SOTA methods on the training sets and test sets of the three scenes of our dataset.
In Table~\ref{traintable} and Table~\ref{testtable}, we show some quantitative results.
The images within the blue borders appear to be clearer and more detailed compared to those within the red borders.
The SOTA methods evaluated on the training set show PSNR values higher than 22, SSIM values around 0.93, and LPIPS values around 0.1, indicating relatively high-quality reconstructions. 
However, when these methods are applied to the test sets, there is a stark contrast in performance.
This indicates that in our dataset, the evaluated methods are severely overfitting on the training samples.

\subsection{Discussion}

The observed overfitting of these methods on our dataset, despite their effective performance on other datasets, can be attributed to several factors: 
1) Our dataset presents complex scenes involving occlusions and interactions, necessitating the concurrent reconstruction of both humans and objects. 
2) Our dataset offers a more extensive range of scenes and exhibits greater variations across different views compared to existing datasets. 
3) The subjects of our dataset are not consistently centralized within the images like other NVS datasets, this may present additional complexities for certain methods.

Our dataset uncovers pronounced overfitting challenges that could impede NVS research involving human-object interactions when applied in real-world scenarios.
This underscores the value of our dataset.
Additionally, we furnish a wealth of prior knowledge within the dataset, including depth sequences, point clouds, foreground masks, mesh files, and SMPL files, intended to assist scholars in mitigating overfitting issues and other challenges.

\begin{table}
\caption{Average Performance of SOTA Methods on The Test Sets of The Three Scenes}
\label{testtable}
\centering
\begin{tabularx}{\linewidth}{m{2.5cm}<{\centering}YYY}
\cmidrule{1-4}
\multirow{2}*{Method} & \multicolumn{3}{c}{Metrics} \\
\cmidrule(l{0.13cm}){2-4}
~ & PSNR$\uparrow$ & SSIM$\uparrow$ & LPIPS$\downarrow$ \\
\cmidrule{1-4}
TensoRF~\cite{chen2022tensorf} & 9.85 & 0.82 & 0.40 \\
K-Planes~\cite{fridovich2023k} & 11.54 & 0.79 & 0.43 \\
3DGS~\cite{kerbl20233d} & 11.22 & 0.51 & 0.28 \\
\cmidrule{1-4}
\end{tabularx}
\end{table}

\section{Conclusion} \label{conclusion}
This paper introduces a new people-object interaction dataset that comprises 38 series of 30-view multi-person or single-person RGB-D video sequences, complemented by corresponding camera parameters, foreground masks, SMPL models, some point clouds, and mesh files.
Each video sequence boasts a 4K resolution, 25 FPS, and a duration of 1$\sim$19 seconds.
All 30 views are captured using Kinect Azure devices in a uniformly surrounding scene.
We also provide NVS benchmarks for our dataset by employing SOTA NVS models.
We hope our work can inspire more research in the study of human-object interaction and NVS.


\vfill\pagebreak

\bibliographystyle{IEEEbib}
\bibliography{refs}

\begin{thebibliography}{10}

\bibitem{wang2023hierarchical}
Shuai Wang, Dehui Kong, Jinghua Li, and Baocai Yin,
\newblock ``Hierarchical hoi detection framework augmented by human interactive
  intention,''
\newblock in {\em 2023 IEEE ICCECT}. IEEE, 2023, pp. 502--508.

\bibitem{su2022robustfusion}
Zhuo Su, Lan Xu, Dawei Zhong, Zhong Li, Fan Deng, Shuxue Quan, and Lu~Fang,
\newblock ``Robustfusion: Robust volumetric performance reconstruction under
  human-object interactions from monocular rgbd stream,''
\newblock {\em IEEE TPAMI}, vol. 45, no. 5, pp. 6196--6213, 2022.

\bibitem{huang2022capturing}
Chun-Hao~P Huang, Hongwei Yi, Markus H{\"o}schle, Matvey Safroshkin, Tsvetelina
  Alexiadis, Senya Polikovsky, Daniel Scharstein, and Michael~J Black,
\newblock ``Capturing and inferring dense full-body human-scene contact,''
\newblock in {\em CVPR}, 2022, pp. 13274--13285.

\bibitem{yoon2021humbi}
Jae~Shin Yoon, Zhixuan Yu, Jaesik Park, and Hyun~Soo Park,
\newblock ``Humbi: A large multiview dataset of human body expressions and
  benchmark challenge,''
\newblock {\em IEEE TPAMI}, vol. 45, no. 1, pp. 623--640, 2021.

\bibitem{andriluka20142d}
Mykhaylo Andriluka, Leonid Pishchulin, Peter Gehler, and Bernt Schiele,
\newblock ``2d human pose estimation: New benchmark and state of the art
  analysis,''
\newblock in {\em CVPR}, 2014, pp. 3686--3693.

\bibitem{chen2015microsoft}
Xinlei Chen, Hao Fang, Tsung-Yi Lin, Ramakrishna Vedantam, Saurabh Gupta, Piotr
  Doll{\'a}r, and C~Lawrence Zitnick,
\newblock ``Microsoft coco captions: Data collection and evaluation server,''
\newblock {\em arXiv preprint arXiv:1504.00325}, 2015.

\bibitem{li2020pastanet}
Yong-Lu Li, Liang Xu, Xinpeng Liu, Xijie Huang, Yue Xu, Shiyi Wang, Hao-Shu
  Fang, Ze~Ma, Mingyang Chen, and Cewu Lu,
\newblock ``Pastanet: Toward human activity knowledge engine,''
\newblock in {\em CVPR}, 2020, pp. 382--391.

\bibitem{goyal2017something}
Raghav Goyal, Samira Ebrahimi~Kahou, Vincent Michalski, Joanna Materzynska,
  Susanne Westphal, Heuna Kim, Valentin Haenel, Ingo Fruend, Peter Yianilos,
  Moritz Mueller-Freitag, et~al.,
\newblock ``The" something something" video database for learning and
  evaluating visual common sense,''
\newblock in {\em ICCV}, 2017, pp. 5842--5850.

\bibitem{jhuang2013towards}
Hueihan Jhuang, Juergen Gall, Silvia Zuffi, Cordelia Schmid, and Michael~J
  Black,
\newblock ``Towards understanding action recognition,''
\newblock in {\em ICCV}, 2013, pp. 3192--3199.

\bibitem{sigurdsson2016hollywood}
Gunnar~A Sigurdsson, G{\"u}l Varol, Xiaolong Wang, Ali Farhadi, Ivan Laptev,
  and Abhinav Gupta,
\newblock ``Hollywood in homes: Crowdsourcing data collection for activity
  understanding,''
\newblock in {\em ECCV}. Springer, 2016, pp. 510--526.

\bibitem{xu2021d3d}
Xiang Xu, Hanbyul Joo, Greg Mori, and Manolis Savva,
\newblock ``D3d-hoi: Dynamic 3d human-object interactions from videos,''
\newblock {\em arXiv preprint arXiv:2108.08420}, 2021.

\bibitem{zhang2022fine}
Lingzhi Zhang, Shenghao Zhou, Simon Stent, and Jianbo Shi,
\newblock ``Fine-grained egocentric hand-object segmentation: Dataset, model,
  and applications,''
\newblock in {\em ECCV}. Springer, 2022, pp. 127--145.

\bibitem{damen2020rescaling}
Dima Damen, Hazel Doughty, Giovanni~Maria Farinella, Antonino Furnari,
  Evangelos Kazakos, Jian Ma, Davide Moltisanti, Jonathan Munro, Toby Perrett,
  Will Price, et~al.,
\newblock ``Rescaling egocentric vision,''
\newblock {\em arXiv preprint arXiv:2006.13256}, 2020.

\bibitem{liu2019ntu}
Jun Liu, Amir Shahroudy, Mauricio Perez, Gang Wang, Ling-Yu Duan, and Alex~C
  Kot,
\newblock ``Ntu rgb+ d 120: A large-scale benchmark for 3d human activity
  understanding,''
\newblock {\em IEEE TPAMI}, vol. 42, no. 10, pp. 2684--2701, 2019.

\bibitem{fan2023arctic}
Zicong Fan, Omid Taheri, Dimitrios Tzionas, Muhammed Kocabas, Manuel Kaufmann,
  Michael~J Black, and Otmar Hilliges,
\newblock ``Arctic: A dataset for dexterous bimanual hand-object
  manipulation,''
\newblock in {\em CVPR}, 2023, pp. 12943--12954.

\bibitem{Kabsch}
Kabsch algorithm,
\newblock ,'' \url{https://en.wikipedia.org/wiki/Kabsch_algorithm}.

\bibitem{lin2021real}
Shanchuan Lin, Andrey Ryabtsev, Soumyadip Sengupta, Brian~L Curless, Steven~M
  Seitz, and Ira Kemelmacher-Shlizerman,
\newblock ``Real-time high-resolution background matting,''
\newblock in {\em CVPR}, 2021, pp. 8762--8771.

\bibitem{zhouibc}
Kai Zhou, Li~Song, Jingchuan Hu, Shuai Guo, Yu~Dong, Yanying Sun, and Yesheng
  Xu,
\newblock ``Realtime 3d reconstruction of dynamic scenes with multiple kinect
  v2 sensors,''
\newblock in {\em IBC}, 2021.

\bibitem{zhou2022rgbd}
Kai Zhou, Shuai Guo, Jingchuan Hu, Jionghao Wang, Qiuwen Wang, and Li~Song,
\newblock ``Rgbd-based real-time volumetric reconstruction system: Architecture
  design and implementation,''
\newblock in {\em 2022 VCIP}. IEEE, 2022, pp. 1--5.

\bibitem{mmhuman3d}
MMHuman3D Contributors,
\newblock ``Openmmlab 3d human parametric model toolbox and benchmark,''
  \url{https://github.com/open-mmlab/mmhuman3d}, 2021.

\bibitem{chen2022tensorf}
Anpei Chen, Zexiang Xu, Andreas Geiger, Jingyi Yu, and Hao Su,
\newblock ``Tensorf: Tensorial radiance fields,''
\newblock in {\em ECCV}. Springer, 2022, pp. 333--350.

\bibitem{fridovich2023k}
Sara Fridovich-Keil, Giacomo Meanti, Frederik~Rahb{\ae}k Warburg, Benjamin
  Recht, and Angjoo Kanazawa,
\newblock ``K-planes: Explicit radiance fields in space, time, and
  appearance,''
\newblock in {\em CVPR}, 2023, pp. 12479--12488.

\bibitem{kerbl20233d}
Bernhard Kerbl, Georgios Kopanas, Thomas Leimk{\"u}hler, and George Drettakis,
\newblock ``3d gaussian splatting for real-time radiance field rendering,''
\newblock {\em ACM ToG}, vol. 42, no. 4, 2023.

\end{thebibliography}


\begin{thebibliography}{1}

\bibitem{chen2022tensorf}
Anpei Chen, Zexiang Xu, Andreas Geiger, Jingyi Yu, and Hao Su,
\newblock ``Tensorf: Tensorial radiance fields,''
\newblock in {\em ECCV}. Springer, 2022, pp. 333--350.

\bibitem{fridovich2023k}
Sara Fridovich-Keil, Giacomo Meanti, Frederik~Rahb{\ae}k Warburg, Benjamin
  Recht, and Angjoo Kanazawa,
\newblock ``K-planes: Explicit radiance fields in space, time, and
  appearance,''
\newblock in {\em CVPR}, 2023, pp. 12479--12488.

\bibitem{kerbl20233d}
Bernhard Kerbl, Georgios Kopanas, Thomas Leimk{\"u}hler, and George Drettakis,
\newblock ``3d gaussian splatting for real-time radiance field rendering,''
\newblock {\em ACM ToG}, vol. 42, no. 4, 2023.

\end{thebibliography}

\end{document}